\documentclass[10pt,twocolumn,letterpaper]{article}

\usepackage{cvpr}              
\usepackage{graphicx}
\usepackage{capt-of}
\usepackage{dblfloatfix}
\usepackage{multirow}
\usepackage{bbding}
\usepackage{placeins} 
\usepackage[table]{xcolor}

\definecolor{cvprblue}{rgb}{0.21,0.49,0.74}
\usepackage[pagebackref,breaklinks,colorlinks,allcolors=cvprblue]{hyperref}
\def\confName{CVPR}
\def\confYear{2026}

\title{DyaDiT: A Multi-Modal Diffusion Transformer for Socially Favorable Dyadic Gesture Generation}

\author{Yichen Peng\textsuperscript{1,2} \quad 
Jyun-Ting Song\textsuperscript{2}\quad 
Siyeol Jung\textsuperscript{2,3} \quad 
Ruofan Liu\textsuperscript{1} \quad 
Haiyang Liu\textsuperscript{4,5} \quad \\
Xuangeng Chu\textsuperscript{4,5} \quad 
Ruicong Liu\textsuperscript{4,5}\quad 
Erwin Wu\textsuperscript{1,4}\quad 
Hideki Koike\textsuperscript{1}\quad
Kris Kitani\textsuperscript{2} 
\\
\\
\textsuperscript{1}Institute of Science Tokyo \quad
\textsuperscript{2}Carnegie Mellon Unversity  \\
\textsuperscript{3}UNIST \quad
\textsuperscript{4}Shanda AI Research Tokyo\quad
\textsuperscript{5}The University of Tokyo \quad
\\
\\
}

\begin{document}

\twocolumn[{%
\maketitle
\centering
\includegraphics[width=\linewidth,height=\textheight,keepaspectratio]{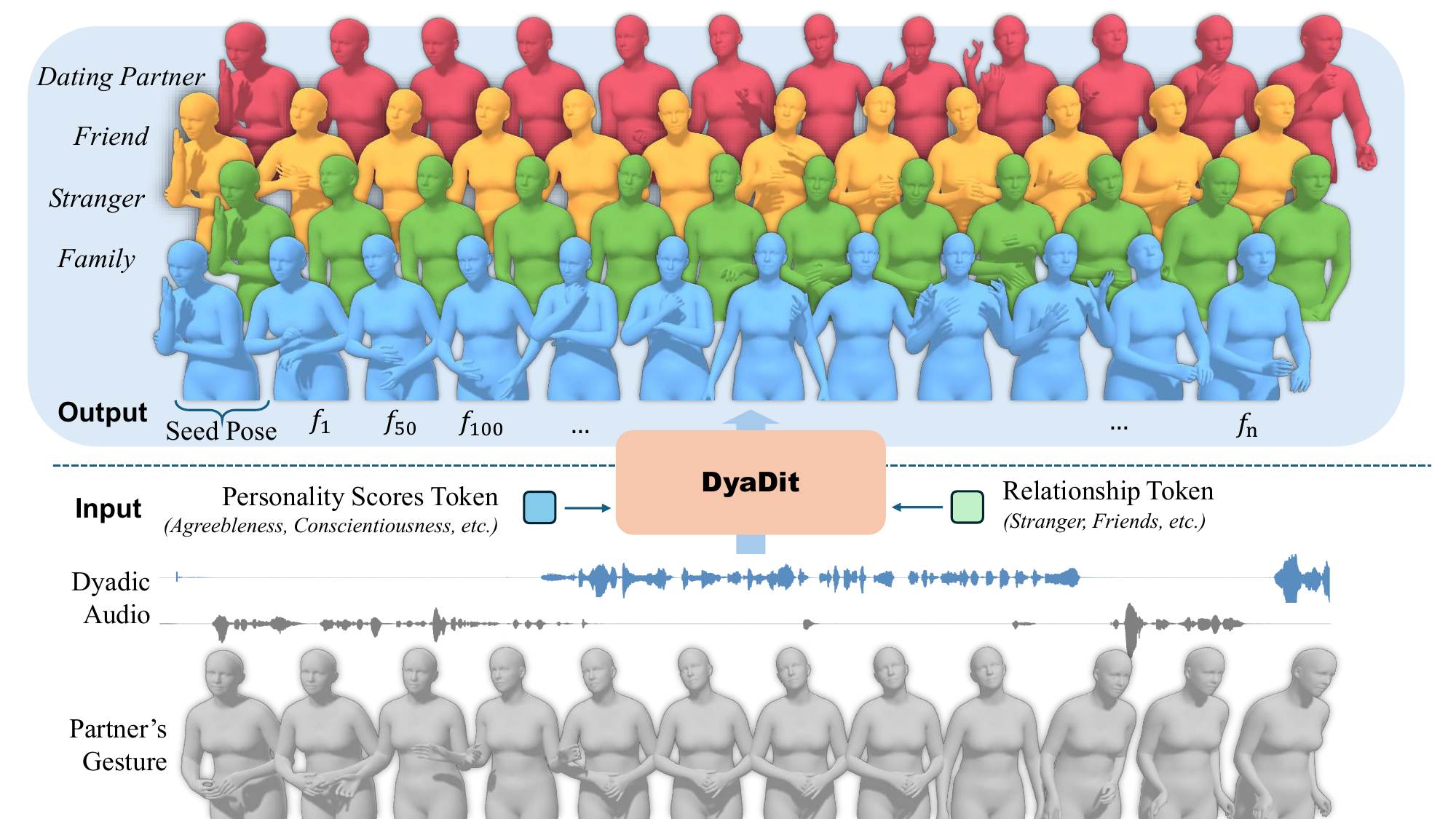}
\captionof{figure}{
\textbf{DyaDiT} generates socially aware conversational gestures from dyadic audio, conditioned on social factors such as relationship and personality traits, achieving natural and contextually appropriate reactions that outperform prior methods in both quantitative and user evaluations.
}
\vspace{0.44cm}
\label{fig:teaser}

}]

\begin{abstract}
Generating realistic conversational gestures are essential for achieving natural, socially engaging interactions with digital humans. However, existing methods typically map a single audio stream to a single speaker’s motion, without considering social context or modeling the mutual dynamics between two people engaging in conversation. We present DyaDiT, a multi-modal diffusion transformer that generates contextually appropriate human motion from dyadic audio signals. Trained on Seamless Interaction Dataset~\cite{seamless_interaction}, DyaDiT takes dyadic audio with optional social-context tokens to produce context-appropriate motion. It fuses information from both speakers to capture interaction dynamics, uses a motion dictionary to encode motion priors, and can optionally utilize the conversational partner's gestures to produce more responsive motion. We evaluate DyaDiT on standard motion generation metrics and conduct quantitative user studies, demonstrating that it not only surpasses existing methods on objective metrics but is also strongly preferred by users, highlighting its robustness and socially favorable motion generation. Code and models can be found at \url{https://puckikk1202.github.io/dyadit_hp/}.

\end{abstract}   
\section{Introduction}
\label{sec:intro}
Building synthetic agents (also known as digital humans, AI agents, avatars or androids) that can interact naturally with people, is essential for the future of human–computer interfaces. Recent language models such as GPT-4.5~\cite{openai2024gpt4technicalreport} and LLaMA-3.1~\cite{grattafiori2024llama3herdmodels} already demonstrate impressive conversational ability, and many users even feel as if they are speaking with another person~\cite{jones2025largelanguagemodelspass}. However, currently this illusion still remains confined to a text window. In reality, human interaction involves more than just spoken words. People gesture, respond to each other, and express subtle social cues through body motion. In order for humans to truly feel that a synthetic agent is interactive, the agent must accompany its speech with gestures that evolve naturally with the conversation.


However, generating such gestures is challenging because they are shaped not only by speech content, but also by rich social context and the dynamics of interaction. Human gestures depend on factors such as personality, the relationship between speakers, and their conversational roles, all of which influence how people move, respond, and coordinate with one another. Yet most existing gesture generation models do not explicitly model these social factors~\cite{tran2024dyadic, reactdiff2025, luo2025omniresponse}, causing the generated motions to appear generic or unnatural. The problem becomes even harder in dyadic conversations, where interaction unfolds dynamically. Two people may speak simultaneously, interrupt one another, or rapidly alternate between speaking and listening~\cite{dunne1994simultaneous,meyer2023timing}. Capturing such moment-to-moment coordination requires the model to distinguish speaker-specific and listener-specific cues, but most existing dyadic gesture generation methods do not explicitly disentangle the two input audio streams~\cite{a2r2024, convofusion2024}, making it difficult for them to generate gestures that faithfully reflect the underlying interaction.

To tackle these limitations, we introduce DyaDiT, a diffusion based transformer that generates socially aware gestures from dyadic audio. Unlike previous work~\cite{emage2024, beat2022, emotiongesture2024, mamba2024} that focuses only on the alignment between audio and generated motion, DyaDiT conditions its generation on explicit social cues such as relationship and personality. Furthermore, in order to address the strong entanglement of two overlapping audio streams during conversation, we propose Orthogonalization Cross Attention (ORCA), a simple yet effective module that disambiguates the two audio streams, resulting in a cleaner audio representation for better gesture generation. Lastly, since human motion in a dyadic setting is often affected by the partner’s gestures, DyaDiT can optionally take the partner’s movements as an additional input, allowing the model to generate gestures that are more coordinated, responsive, and natural.

Our experiments show that DyaDiT consistently outperforms existing dyadic gesture generation methods across standard quantitative metrics, achieving clear improvements in gesture quality and distribution alignment. In addition, we conduct extensive user studies to assess human perceptual preference. The results show that participants strongly favor gestures generated by DyaDiT, indicating that our generated motion appears more socially aware and better suited for real conversational settings.

In summary, our main contributions are as follows:
\begin{itemize}
\item We propose DyaDiT, a diffusion transformer (DiT) that generates social context aware gestures in dyadic conversations.
\item We introduce an Orthogonalization Cross Attention (ORCA) module that reduces interference between two individual's audio streams, enabling a cleaner audio representation for better gesture generation in dyadic conversation.
\item Through extensive quantitative evaluations and user studies, we show that DyaDiT consistently outperforms existing methods on standard metrics and is preferred by users in terms of perceived realism and social consistency.
\end{itemize}

\begin{figure*}[!t]
    \centering
    \includegraphics[width=\linewidth]{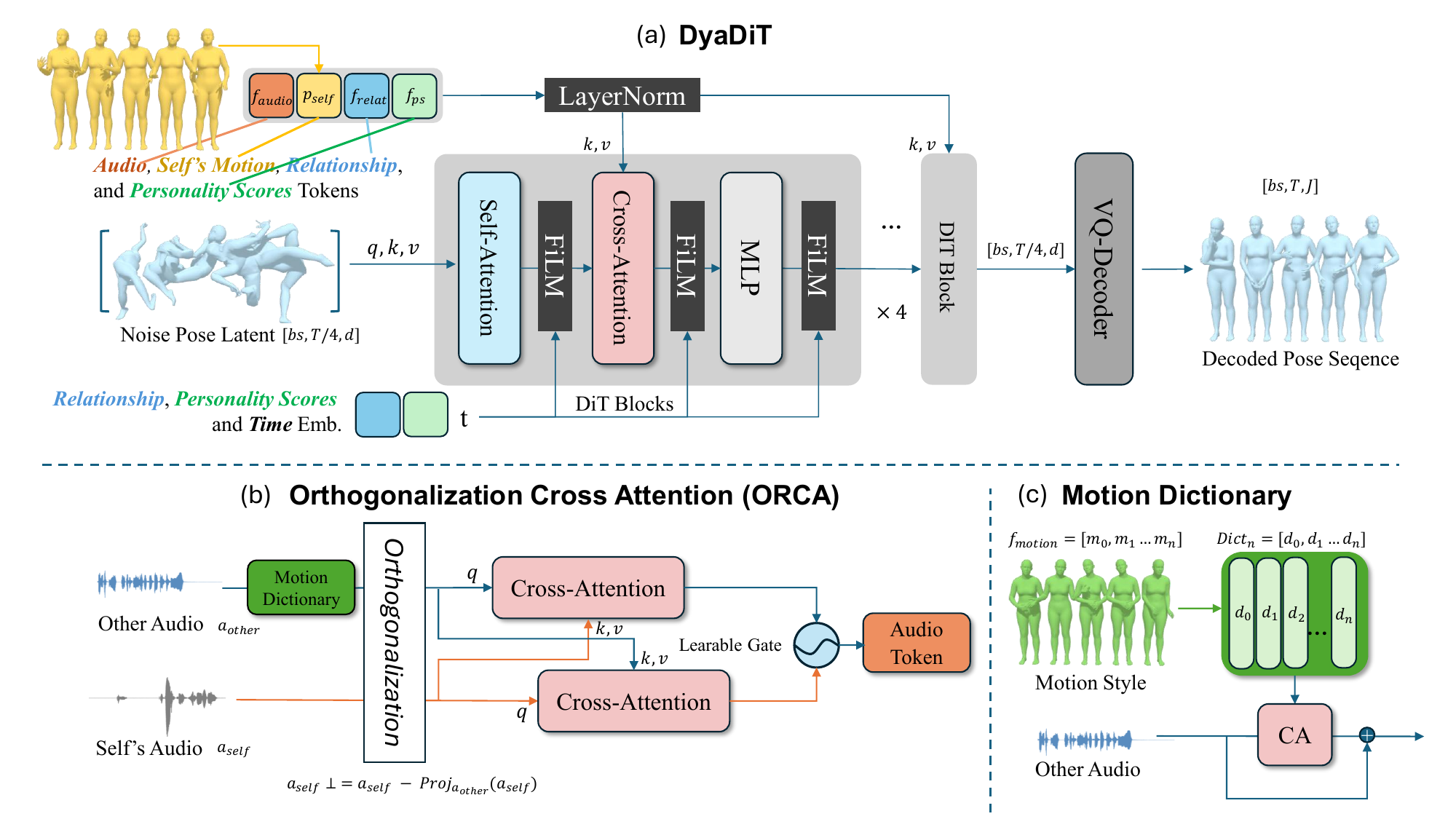}
    \caption{\textbf{Overview of DyaDiT.} DyaDiT conditions on multiple input modalities, including audio, partner motion, relationship type, and personality scores. It employs an Audio Orthogonalization Cross Attention (ORCA) module to obtain cleaner audio representations and a motion dictionary to guide style aware gesture generation.} 
    \label{fig:framework}
\end{figure*}

\section{Related Work}
\label{sec:relatedwork}
\subsection{Co-Speech Gesture Generation}

Co-speech gesture generation focuses on synthesizing body movements aligned with a single speaker's speech. Early approaches formulate gesture generation as a translation problem from multimodal speech cues, combining text, audio, and speaker identity through recurrent or adversarial models, such as the trimodal framework of Yoon et al.~\cite{yoon2020}, which treated gesture generation as a translation problem, combining speech transcripts, audio, and speaker identity through an adversarial recurrent framework. Liu et al.~\cite{beat2022} introduced the BEAT dataset and CaMN, a cascaded multimodal adversarial network capable of generating synchronized body and hand gestures. Other works, such as EMAGE~\cite{emage2024}, TalkSHOW~\cite{yi2023talkshow}, MECo~\cite{meco2025}, etc.~\cite{bodyformer2023, liu2022learning, S2G2019, disco2022, yang2024Freetalker} enhanced realism by disentangling rhythmic and semantic motion features or integrating discrete latent spaces via VQ-VAEs.

Although these advances are notable, most of them focuses on single-speaker co-speech gestures and overlooks the interactive nature of human communication. More complex settings such as dyadic gesture generation, which require modeling both participants' behaviors and their interpersonal dynamics, remain largely unexplored.

\subsection{Dyadic Gesture and Reaction Generation}
Unlike co-speech gesture synthesis, dyadic gesture generation must model the coordinated behavior between two participants, including interpersonal timing, mutual attention, and responsiveness. Most prior works in this domain stem from facial reaction generation (FRG), where the goal is to predict non-deterministic listener facial responses to a speaker’s behavior~\cite{apb2face2020,10890674,reactdiff2025,luo2025omniresponse,reactface2025,song2025react,Ng2022face}. Beyond FRG, related efforts have also been made in broader interactive motion domains, such as couple dance modeling~\cite{maluleke2024synergysynchronycoupledances} and two-person motion dataset \& synthesis~\cite{khirodkar2024harmony4d, remos2024, siyao2024duolando}, both of which emphasize the role of interpersonal synchrony and response modeling.
Although FRG provides a basis for modeling two-person interactions, it primarily focuses only on facial or head movements rather than full-body gestures. Due to this limitation, several recent works have begun exploring body gesture generation in dyadic settings.
A few efforts such as Audio2Photoreal~\cite{a2r2024}, ConvoFusion~\cite{convofusion2024}, and TAG2G~\cite{tag2g2024} extend single-speaker frameworks to two-party settings. 
Yet, these models often either (1) overlook the social context between the two individuals or (2) treat dyadic audio as a single blended signal without explicitly modeling cross-person dynamics, which often leads to ambiguity in the roles and interaction patterns presented to the model. These limitations highlight the need for explicit feature disentanglement and better support for socially contextual reasoning.


\subsection{Diffusion-based Gesture Generation}
Diffusion models have emerged as powerful generative tools for human motion due to their ability to model multi-modal and many-to-many distributions.
Several works have adapted diffusion frameworks originally designed for text-conditioned motion generation, such as MotionDiffuse~\cite{motiondiffusion2024}, FineMoGen~\cite{zhang2023finemogen}, and MDM~\cite{tevet2023mdm}, etc.~\cite{shafir2024human, tevet2022motionclip, chen2023executing, Ao2023GestureDiffuCLIP}, to the speech–gesture domain. Alexanderson et al.~\cite{alexanderson2023} reformulated DiffWave for co-speech gestures, while Zhu et al.~\cite{zhu2023diffgesture} proposed DiffGesture, integrating noisy gesture sequences with contextual embeddings for temporal modeling. DiffuseStyleGesture+~\cite{diffstylegesture2023} further introduced conditioning on audio, text, style, and seed gestures, leveraging transformer-based denoising with attention control. Other works, such as UnifiedGesture~\cite{unifiedgesture2023}, LivelySpeaker~\cite{livespeaker2023}, and AMUSE~\cite{amuse2024}, emphasized semantic and rhythmic consistency, disentangling emotional and stylistic latent factors. Considering the success of diffusion-based approaches in gesture generation, and the inherently nondeterministic nature of dyadic conversation, we build our method upon a diffusion model to better capture the variability and dynamics present in two-person interactions.






\begin{figure*}[th!]
    \centering
    \includegraphics[width=\linewidth]{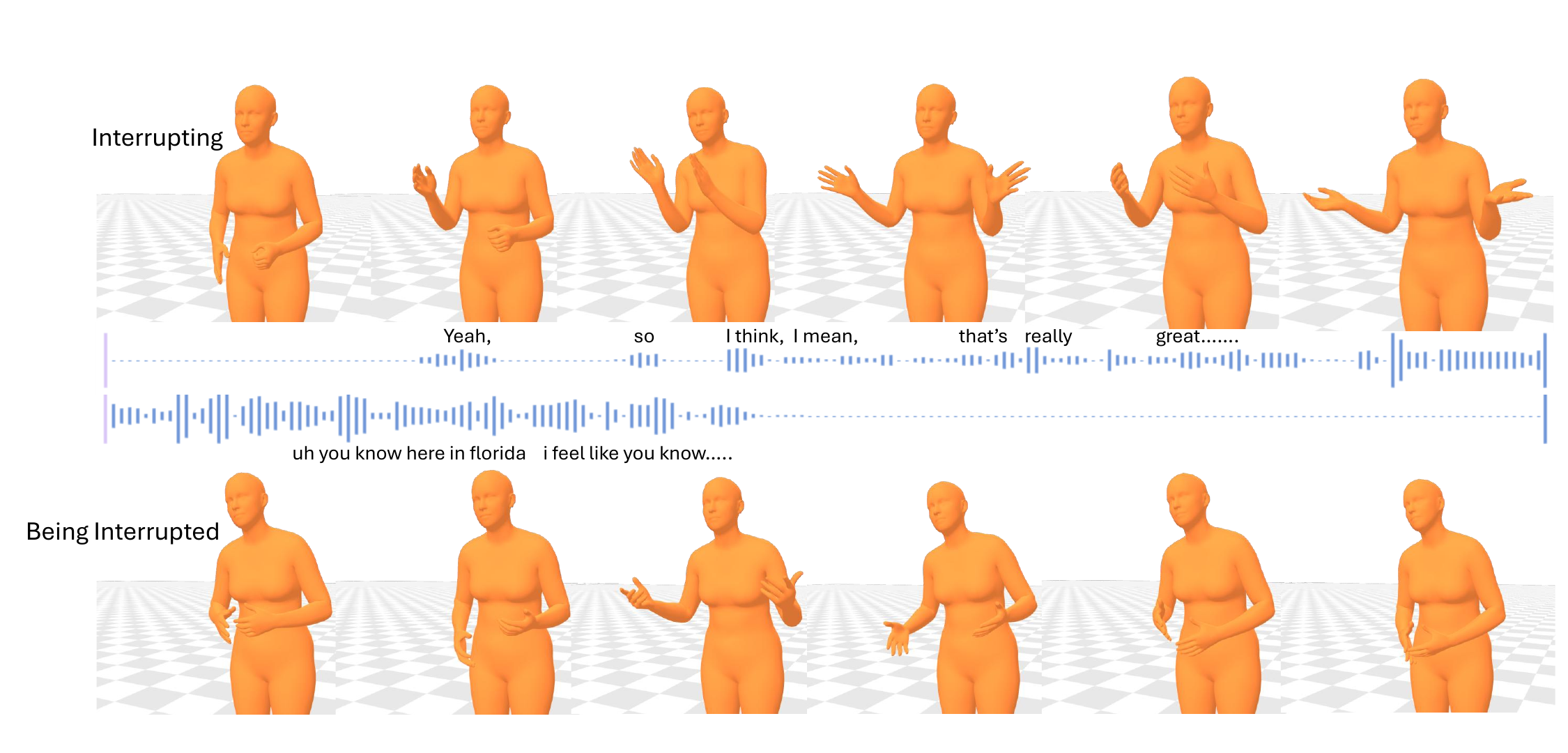}
    \caption{\textbf{ORCA} reduces ambiguity between the two audio streams, allowing \textbf{DyaDiT} to generate realistic motion even when one person interrupts the other during the conversation. The example demonstrates the generated motions adjusts naturally as the conversation shifts.}
    \label{fig:interrupt}
\end{figure*}

\section{DyaDiT}

We introduce \textbf{DyaDiT}, a diffusion transformer (DiT) with multi-modal input for dyadic gesture generation. Given the conversational audio streams from two individuals (denoted as \textit{self} and \textit{other}), our objective is to generate plausible upper-body gestures for the \textit{other} speaker in accordance with the conversational context. To further enhance the realism of the generated gestures, DyaDiT optionally incorporates the relationship, personality traits, and gesture sequence of \textit{self} as auxiliary conditioning signals during inference. We train our model on a subset of the Seamless Interaction dataset~\cite{seamless_interaction}, with dataset specifications and preprocessing procedures detailed in Section~\ref{sec:dataset}. An overview of the full architecture is shown in Fig.~\ref{fig:framework}. Details of the Audio Orthogonalization Cross Attention (ORCA) module are provided in Section~\ref{sec:bidirectional}, while the motion dictionary and motion tokenizer are presented in Sections~\ref{sec:dictionary} and ~\ref{sec:VQ-VAE}, respectively.

\subsection{Seamless Interaction Dataset}
\label{sec:dataset}
The Seamless Interaction dataset~\cite{seamless_interaction} is a large-scale corpus of dyadic conversation containing synchronized audio, full-body motion, and facial expressions. In this work, we use a curated subset of 3{,}000 clips (approximately 182 hours) from the dataset’s \textit{naturalistic} scenario collection, which contains rich and spontaneous dyadic conversations suited for modeling conversational gestures. We focus on generating upper-body gestures, represented as $g \in \mathbb{R}^{T \times J \times 6}$ using the 6D rotation representation~\cite{zhou2018rot6d}, where $J$ denotes the number of upper-body joints. In addition to motion data, we incorporate two high level social annotations provided in the dataset: a relationship type label \( f_{rs} \in \{0,1\}^{4} \) indicating whether the speakers are friends, strangers, family members, or dating partners; and a personality score vector \( f_{ps} \in \mathbb{R}^{5} \) that quantifies five major personality traits, which are extraversion, agreeableness, conscientiousness, neuroticism, and openness. For audio inputs, we extract the dyadic speech signals from both individuals and process them with a pretrained Wav2Vec2 encoder~\cite{wav2vec2} to obtain audio feature embeddings for conditioning in our model.

\subsection{DiT Backbone}
\label{sec:backbone}

Our model adopts a diffusion transformer (DiT) backbone that follows the Denoising Diffusion Probabilistic Model (DDPM)~\cite{ho2020denoising} framework. The network takes a noisy latent pose $\mathbf{x}_t$ and predicts the added Gaussian noise according to time step $t$, $\boldsymbol{\epsilon}_\theta(\mathbf{x}_t, t, \mathbf{c})$ conditioned on multiple contextual inputs $\mathbf{c}$, 
where $\mathbf{c} =(ORCA(a_{\text{self}}, a_{\text{other}}), p_{\text{self}}, f_{\text{relat}}, f_{\text{ps}})$ including audio, partner’s motion, relationship type, and personality scores. The training objective follows the standard $\epsilon$-prediction loss:
\begin{equation}
\mathcal{L}_{\text{diff}} = 
\mathbb{E}_{\mathbf{x}_0, t, \boldsymbol{\epsilon}} 
\left[
    \left\| 
        \boldsymbol{\epsilon} - 
        \boldsymbol{\epsilon}_\theta(\mathbf{x}_t, t, \mathbf{c}) 
    \right\|_2^2
\right],
\end{equation}
where $\mathbf{x}_t = \sqrt{\bar{\alpha}_t}\mathbf{x}_0 + \sqrt{1-\bar{\alpha}_t}\boldsymbol{\epsilon}$.

As shown in Figure~\ref{fig:framework}(a), each DiT block consists of a self-attention layer for modeling temporal dependencies within the latent pose sequence and a cross-attention layer for integrating contextual information from multimodal cues. In addition, the relationship and personality embeddings are injected through both FiLM-based~\cite{film2018} modulation and cross-attention, enabling DyaDiT to jointly capture social attributes and individual expressive styles in gesture generation.

\subsection{Audio Orthogonalization Cross Attention (ORCA)}
\label{sec:bidirectional}

To effectively capture conversational dynamics between two speakers, we introduce a \textbf{orthogonalization cross attention} module for audio fusion (Figure~\ref{fig:framework}(b)). Given the audio features from both speakers $a_{\text{self}}$ and $a_{\text{other}}$, encoded by Wav2Vec2, our goal is to separate redundant factors while aligning complementary information into a joint representation.

We first apply an \textit{orthogonalization} process to filter out redundant components between the two audio streams:
\begin{equation}
a_{\text{self}}^{\perp} = a_{\text{self}} - \mathrm{Proj}_{a_{\text{other}}}(a_{\text{self}}),
\end{equation}
where $\mathrm{Proj}_{a_{\text{other}}}(\cdot)$ denotes the projection of the self audio feature $a_{self}$ onto the subspace of the other’s audio feature, implemented as a lightweight MLP: 
\begin{equation}
\phi(\mathbf{x}) = W_2 \sigma(W_1 \mathbf{x} + b_1) + b_2.
\end{equation}
This operation enforces complementary conditioning and reduces correlated information across dyadic audio.

Subsequently, we employ two symmetric cross-attention modules to achieve bidirectional information exchange. The first module uses $a_{\text{other}}$ as query and attends to $a_{\text{self}}^{\perp}$, capturing the speaker’s response to the partner’s utterance. Conversely, the second module takes $a_{\text{self}}^{\perp}$ as query and attends to $a_{\text{other}}$, modeling the listener’s reactive cues. The outputs of the two cross-attention streams are then adaptively fused through a learnable gating mechanism:
\begin{equation}
f_{\text{audio}} = \sigma(\mathbf{W}_g) \cdot h_{\text{self}\rightarrow\text{other}} + 
(1 - \sigma(\mathbf{W}_g)) \cdot h_{\text{other}\rightarrow\text{self}},
\end{equation}
where $\sigma(\cdot)$ is a sigmoid function and $\mathbf{W}_g$ is a learnable gate parameter.

The resulting fused token $f_{\text{audio}}$ is used as the final audio conditioning input to the DyaDiT, providing a plausible acoustic embedding that reflects both interlocutors’ vocal behaviors. As shown in Figure~\ref{fig:interrupt}, with ORCA, DyaDiT have the capability to generate either speaker or listener gestures.

\subsection{Motion Dictionary (MD)}
\label{sec:dictionary}
Inspired by prior work LIA~\cite{wang2024lia} which learns a set of orthogonal motion-directions by enforcing orthonormality at initialization, we similarly introduce a learnable orthogonal motion dictionary to modulate the partner’s audio feature $a_{other}$ according to the current style motion. The motion dictionary is a module designed to incorporate motion style conditioning when user requires motion style control. As shown in Figure~\ref{fig:framework}(c), it consists of a set of learnable motion bases $\{d_0, d_1, \dots, d_n\}$ that encode representative gesture primitives. During training, we include ground truth motion style features $f_{\text{motion}} = [m_0, m_1, \dots, m_n]$ to guide the model in learning style-aware correspondences between audio cues and motion patterns. 

Given the other-speaker audio feature $a_{\text{other}}$, the dictionary is integrated through a cross-attention (CA) operation followed by a weighted combination with the motion bases:

\begin{equation}
a_{\text{other}}' = \mathrm{CA}(a_{\text{other}}, \sum_{k=0}^{n} m_k d_k) + a_{\text{other}},
\end{equation}

where $\mathrm{Dict} = [d_0, d_1, \dots, d_n]$ and $m_k$ denotes the style-dependent modulation weight derived from $f_{\text{motion}}$. 

The Motion Dictionary is jointly trained with DyaDiT without orthogonalization, as strict phase alignment is unnecessary for gestures and degrades learning. At inference, the motion dictionary can be optionally activated. With classifier-free guidance (CFG)~\cite{ho2021classifierfree}, the style conditioning can be strengthened to generate gestures with distinct stylistic patterns, or completely dropped to produce an unconditional, style-agnostic motion driven purely by the learned audio–motion prior.

\subsection{Motion Tokenizer}
\label{sec:VQ-VAE}
Instead of directly training our DiT model using the raw gesture sequences $x_0' \in \mathbb{R}^{T \times J}$, we first train a VQ-VAE to discretize the continuous motion space into compact latent representations. Our implementation follows the residual vector quantization framework introduced in prior works~\cite{lee2022autoregressiveimagegenerationusing}, where multiple codebooks are cascaded to progressively refine quantization errors. Specifically, we adopt a residual VQ-VAE with a residual length of 4, utilizing four hierarchical codebooks to capture progressively finer-grained motion details. In addition, to reduce temporal redundancy, we downsample the input gesture sequence by a factor of 4 using 1D convolution layers in the encoder, resulting in one latent token for every four frames. During training of DyaDiT, the VQ-VAE encoder compresses gesture sequences into a sequence of quantized tokens, $x_t' \in \mathbb{R}^{T/4 \times d}$, where $d=64$ in our experiment. At inference time, the VQ-VAE decoder then reconstructs continuous motions from the generated quantized embeddings. Note that since our diffusion model operates directly in the latent space, the encoder is not used during inference, and the decoder is not used during training. This hierarchical and temporally compact quantization allows the diffusion model to capture long-range dependencies while maintaining high-fidelity motion reconstruction.

\section{Experiments}

We evaluate the effectiveness of DyaDiT for dyadic gesture generation through both quantitative and qualitative experiments. For the quantitative evaluation, we compute the Fréchet Distance (FD) and Diversity metrics to compare DyaDiT with two existing dyadic gesture generation models, ConvoFusion~\cite{convofusion2024} and Audio2PhotoReal~\cite{a2r2024}. In addition, we report scores with respect to the ground truth gestures, which serve as a reference for the underlying motion distribution in the dataset. For the qualitative evaluation, we conduct a user preference study to evaluate both the overall perceived quality of the generated gestures and their alignment with the social relationship and personality characteristics.
\subsection{Evaluation Metrics}
\vspace{-0.2cm}
\begin{table}[t]
\centering
\fontsize{7}{8}\selectfont
\newcolumntype{R}{c}
\begin{tabular}{l|R|cc|cc}
\toprule
\multirow{2}{*}{Method}
& \multirow{2}{*}{BC}
& \multicolumn{2}{c|}{FD}
& \multicolumn{2}{c}{Diversity} \\
\cmidrule[\cmidrulewidth](lr){3-4}\cmidrule[\cmidrulewidth](lr){5-6}
& ($\times10^{-1}$) & Sta. $\downarrow$ & Kine. $\downarrow$
& Sta. $\uparrow$ & Kine. $\uparrow$ \\
\midrule
\rowcolor{gray!10} GT        
& - &  -    &  -   & 28.42 & 1.97 \\
\rowcolor{gray!10} GT Random 
& - & 14.94 & 3.74 & 33.85 & 2.05 \\
\midrule
ConvoFusion 
& - & 9.22 & 1.74 & 18.33 & 1.10 \\
Audio2PhotoReal
& - & 8.77 & 1.84 & 19.35 & 1.05 \\
\midrule
DyaDiT (w/o ORCA) 
& 7.25 & 7.32 & 1.79 & 23.57 & 1.24 \\
DyaDiT (CrossAttn) 
& 7.10 & 7.82 & 1.91 & 18.87 & 1.13 \\
DyaDiT (w/o MD) 
& 7.37 & 6.88 & 1.75 & 18.34 & 1.29 \\
DyaDit (MD contin) 
& 7.39 & 6.69 & 1.72 & 21.47 & 1.33 \\
DyaDiT (w/o self) 
& 7.61 & 6.64 & 1.48 & 26.43 & 1.25 \\
DyaDiT (Uncond) 
& 7.63 & 7.40 & 1.63 & 21.65 & 1.16 \\
DyaDiT (Random) 
& 7.64 & 8.24 & 1.53 & 21.94 & \cellcolor{green!15} \textbf{1.43} \\
DyaDiT 
& \cellcolor{green!15} \textbf{7.71} & \cellcolor{green!15} \textbf{6.40} & \cellcolor{green!15} \textbf{1.37} & \cellcolor{green!15} \textbf{27.46}& 1.38 \\
\bottomrule
\end{tabular}
\caption{Quantitative comparison of DyaDiT and baselines in terms of Fréchet Distance (FD) and Diversity. Lower FD indicates higher realism, and higher diversity values indicate more varied motion generation.}
\label{tab:quantitative}
\vspace{-0.7cm}
\end{table}

\begin{figure*}[!t]
    \centering
    \includegraphics[width=\linewidth]{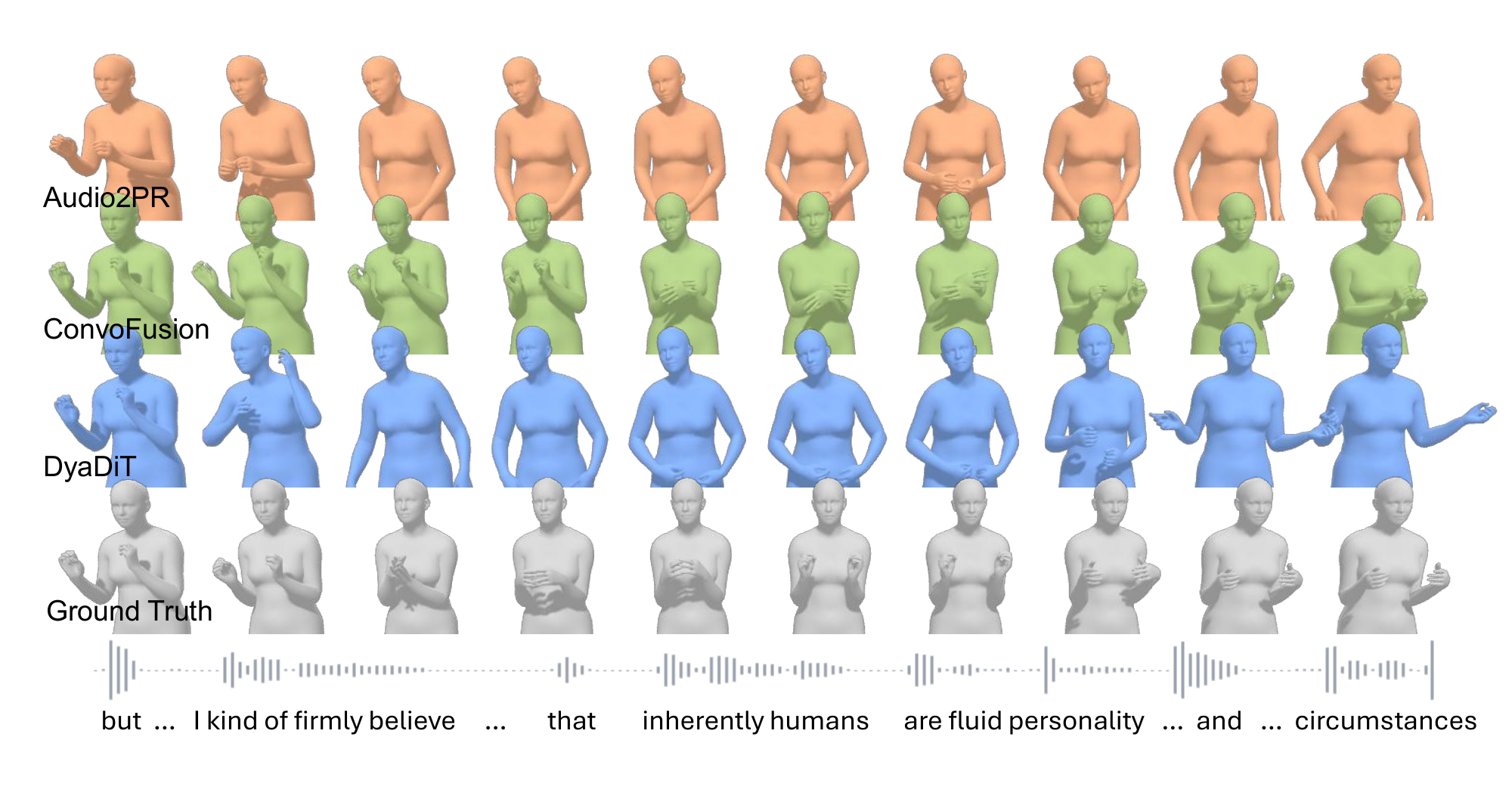}
    \vspace{-1cm}
    \caption{\textbf{Qualitative Results.} Comparison of visualization results between \textbf{DyaDiT}, \textbf{ConvoFusion}~\cite{convofusion2024}, and \textbf{Audio2PhotoReal}~\cite{Ao2023GestureDiffuCLIP}. The gestures generated by DyaDiT exhibit higher diversity and greater realism compared to the other methods.}
    \label{fig:comparison}
    \vspace{0.3cm}
\end{figure*}

\begin{figure}[!t]
    \centering
    \hspace{2cm}
    \includegraphics[width=\linewidth]{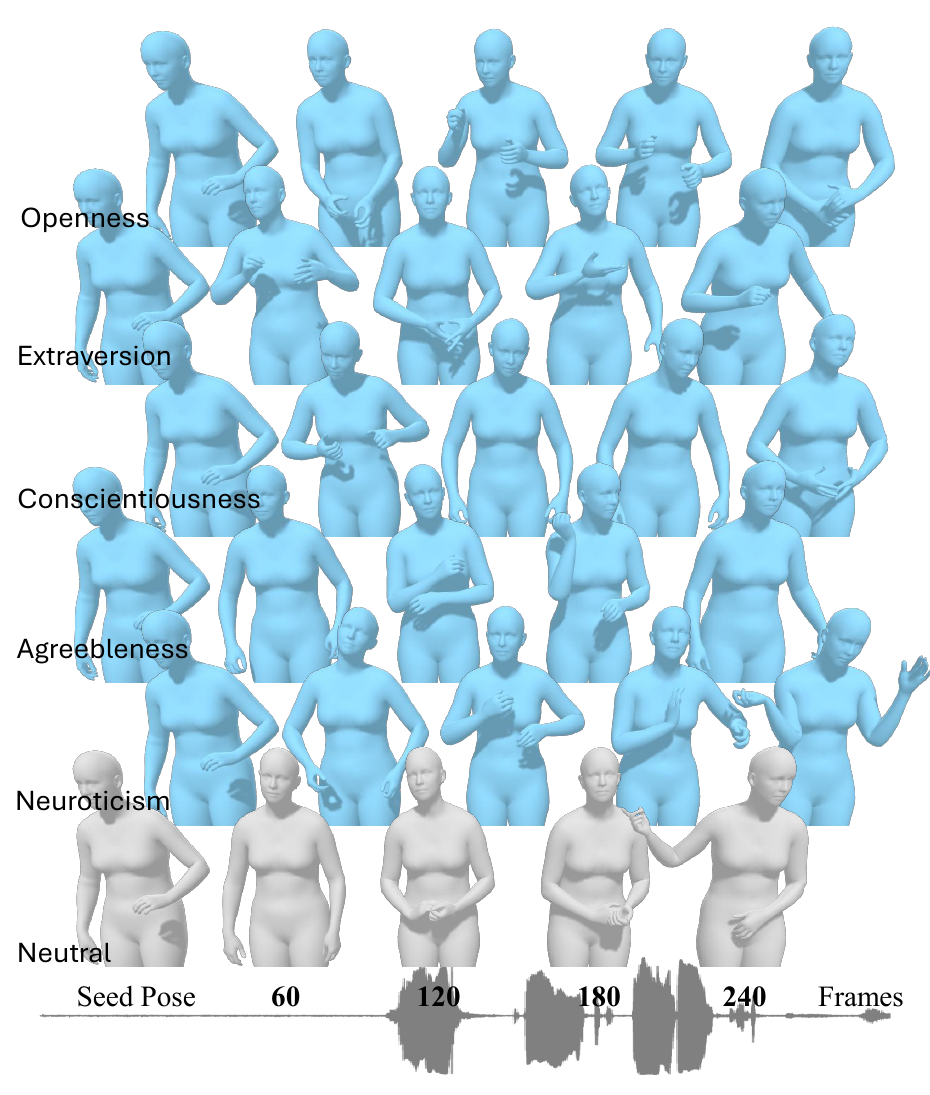}
    \caption{Visualization results under different personality score conditionings. All samples are generated using classifier-free guidance with CFG = 2.5.}
    \label{fig:personality}
\end{figure}

Following prior works on gesture generation~\cite{a2r2024,Ao2023GestureDiffuCLIP,alexanderson2023}, we adopt both distribution and diversity metrics for quantitative evaluation. Specifically, we report the \textit{Fréchet Distance} (FD) to measure the realism of generated motions, \textit{Diversity}~\cite{audio2gesture} to quantify motion variability across samples
For reference, we also compute all metrics on ground truth samples from the Seamless Interaction dataset\cite{seamless_interaction} to provide a baseline for comparison. Below, we provide the definitions of the quantitative metrics used in our evaluation.

\begin{itemize}
    \item \textbf{BC} measures beat consistency between generated gestures and audio, reflecting how well motion accents are synchronized with speech beats.
    \item \textbf{FD (Static)} measures the pose realism for individual frames in the pose space $\mathbb{R}^{3\times j}$, where $j{=}43$ (including finger joints). 
    \item \textbf{FD (Kinetic)} evaluates the realism of temporal dynamics by computing the FD over gesture velocities in $\mathbb{R}^{T\times 3\times j}$, where $T{=}300$ frames in our experiments.
    \item \textbf{Diversity (Static)} quantifies the variation of single-frame poses by computing the average mean squared error (MSE) between clips.
    \item \textbf{Diversity (Kinetic)} measures temporal motion diversity by averaging the pairwise velocity differences across generated gesture sequences.
\end{itemize}

\subsection{Baseline Methods}


We compare DyaDiT with two representative dyadic gesture generation baselines: \textbf{ConvoFusion}~\cite{convofusion2024}, a diffusion-based multimodal fusion model with stacked cross-attention layers, and \textbf{Audio2PhotoReal}~\cite{a2r2024}, which predicts gesture keyframes from audio and refines them with a diffusion transformer. For a fair comparison, both models are adapted to our gesture representation, retrained on the same training subset of the Seamless Interaction Dataset~\cite{seamless_interaction}, and conditioned on the same seed pose. During inference, both methods use a DDIM scheduler with 50 denoising steps (CFG = 2) to generate 300-frame (10\,s) gesture sequences.

\subsection{Quantitative Results}


The quantitative comparison between \textbf{DyaDiT} and baseline models is shown in Table~\ref{tab:quantitative}. For reference, \textbf{GT} diversity reflects the upper bound of motion variation in the dataset, while \textbf{GT Random} \textit{FD} provides a lower-bound reference for cross-sample mismatch. Overall, \textbf{DyaDiT} achieves the best overall performance while maintaining strong motion diversity. The ablation results further confirm the contributions of ORCA, the discrete Motion Dictionary, and self-conditioning, each of which improves generation quality and interaction modeling. These results demonstrate the effectiveness of the proposed social context-aware DiT framework over prior baselines such as ConvoFusion~\cite{convofusion2024} and Audio2PhotoReal~\cite{audio2gesture}.

\subsection{Ablation Studies}
We further conduct several ablation studies to analyze how different components and conditioning signals contribute to DyaDiT’s performance. Below, we explicitly explain the difference between each variant. 
\begin{itemize}
    \item \textbf{DyaDiT (w/o ORCA)} removes the ORCA module and directly concatenates the raw dyadic audio features.
    \item \textbf{DyaDiT (CrossAttn)} replaces the ORCA module with a standard cross-attention for combining \textit{self} and \textit{other} audio features.
    \item \textbf{DyaDiT (w/o MD)} removes the motion dictionary.
    \item \textbf{DyaDiT (MD contin)} replaces the discrete motion dictionary with a continuous motion representation.
    \item \textbf{DyaDiT (w/o self)} removes the \textit{self} motion conditioning branch.
    \item \textbf{DyaDiT (Uncond)} generates gestures without any conditioning.
    \item \textbf{DyaDiT (Random)} performs inference with randomly assigned relationship and personality labels.
\end{itemize}

As expected, removing \textbf{ORCA} degrades performance, especially in \textit{FD}, confirming its role in improving motion realism. Replacing the discrete \textbf{Motion Dictionary} with a continuous variant also hurts performance, particularly in \textit{Diversity (Static)}, indicating that discrete motion bases better capture diverse interaction styles.
We also find that the full \textbf{DyaDiT} outperforms both \textbf{Uncond} and \textbf{Random} social conditioning in overall generation quality, showing the importance of social context for realistic gesture generation. In particular, removing or mismatching social context noticeably reduces \textit{Diversity (Static)}, suggesting that proper social cues help the model produce a broader motion distribution.
Finally, \textbf{DyaDiT (Uncond)} does not achieve higher diversity than the fully conditioned model. We attribute this to the dyadic setting, where many segments correspond to listener behavior with inherently limited motion variation; additional social conditioning helps disambiguate such cases and encourages richer gestures.



\section{User Study}
\begin{figure}
    \centering
    \includegraphics[width=1.05\linewidth]{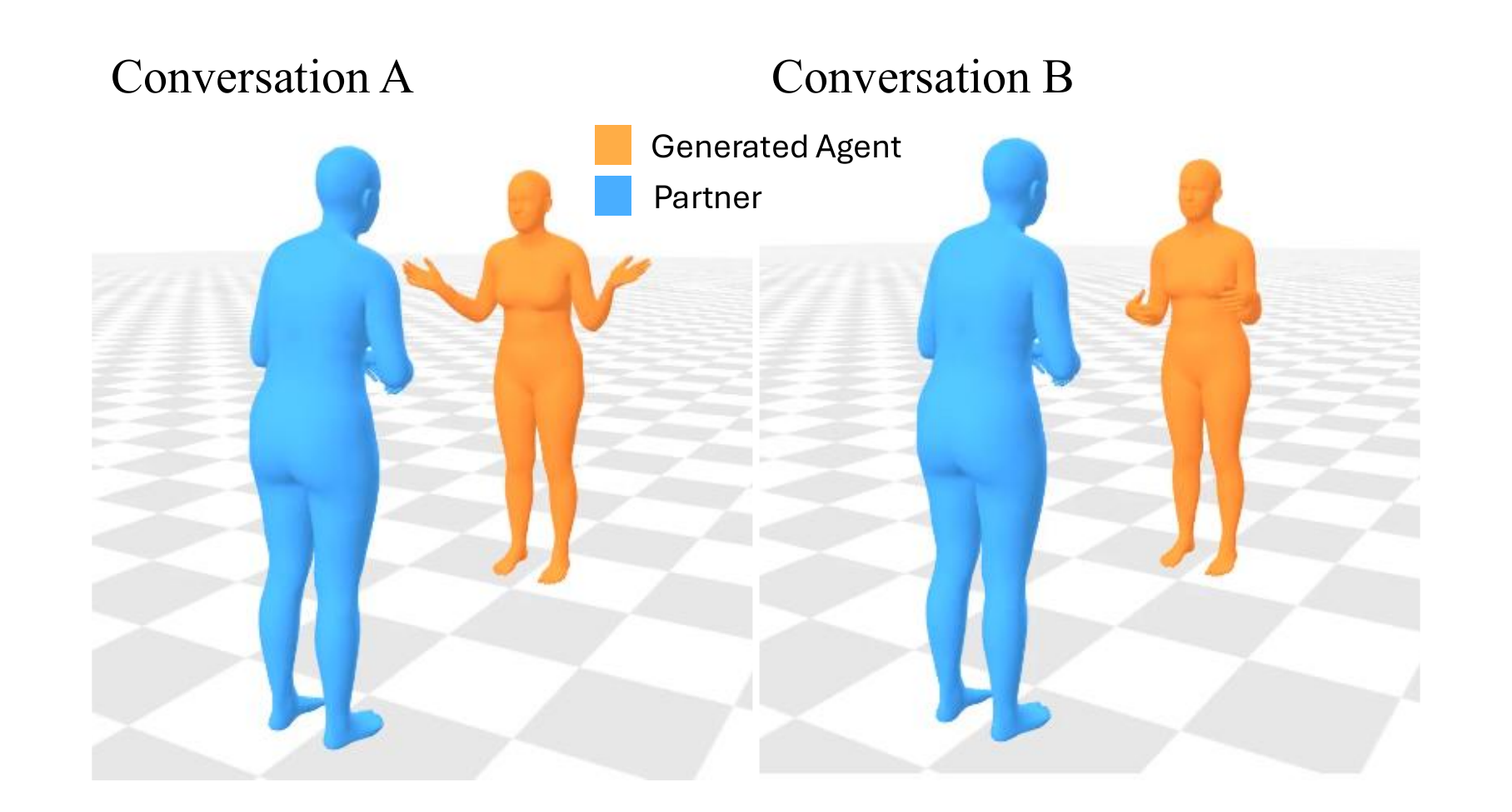}
    \caption{Example video pairs used in the user study for evaluating participant preference in conversational gesture generation.}
    \label{fig:userstudysample}
\end{figure}

We conduct an A/B preference study with sixteen participants to assess user preference on human motion in dyadic conversation settings. We compare our method with ConvoFusion~\cite{convofusion2024} as well as with ground truth motion. The user study evaluates three aspects of the generated gestures: overall quality, relationship consistency, and personality consistency. Participants are shown paired clips and select their preferred gesture sequence, with the video pairs presented in randomized order to avoid ordering bias. An example video pair is shown in Figure~\ref{fig:userstudysample}. 

We randomly selected 56 ten-second sequences from the validation set of the Seamless Interaction dataset to conduct the experiment. Below we show sample questions presented to the participants in each section.
\begin{itemize}
\item \textbf{overall quality}: \textit{“Which gesture of the orange character looks more human-like?”}
\item \textbf{relationship consistency}: \textit{“Which pair seems more likely to be friends (or strangers, or family members, or dating partners)?”}
\item \textbf{personality consistency}: \textit{“Which gesture better reflects the personality trait agreeable (or conscientious, or extraverted, or neurotic)?”}
\end{itemize}

For each video pair, participants were asked to provide both their preference and their confidence level (strongly or slightly). For the full questionnaire and interface design used in the experiment, please refer to the supplementary material.
\begin{figure}
    \centering
    \includegraphics[width=\linewidth]{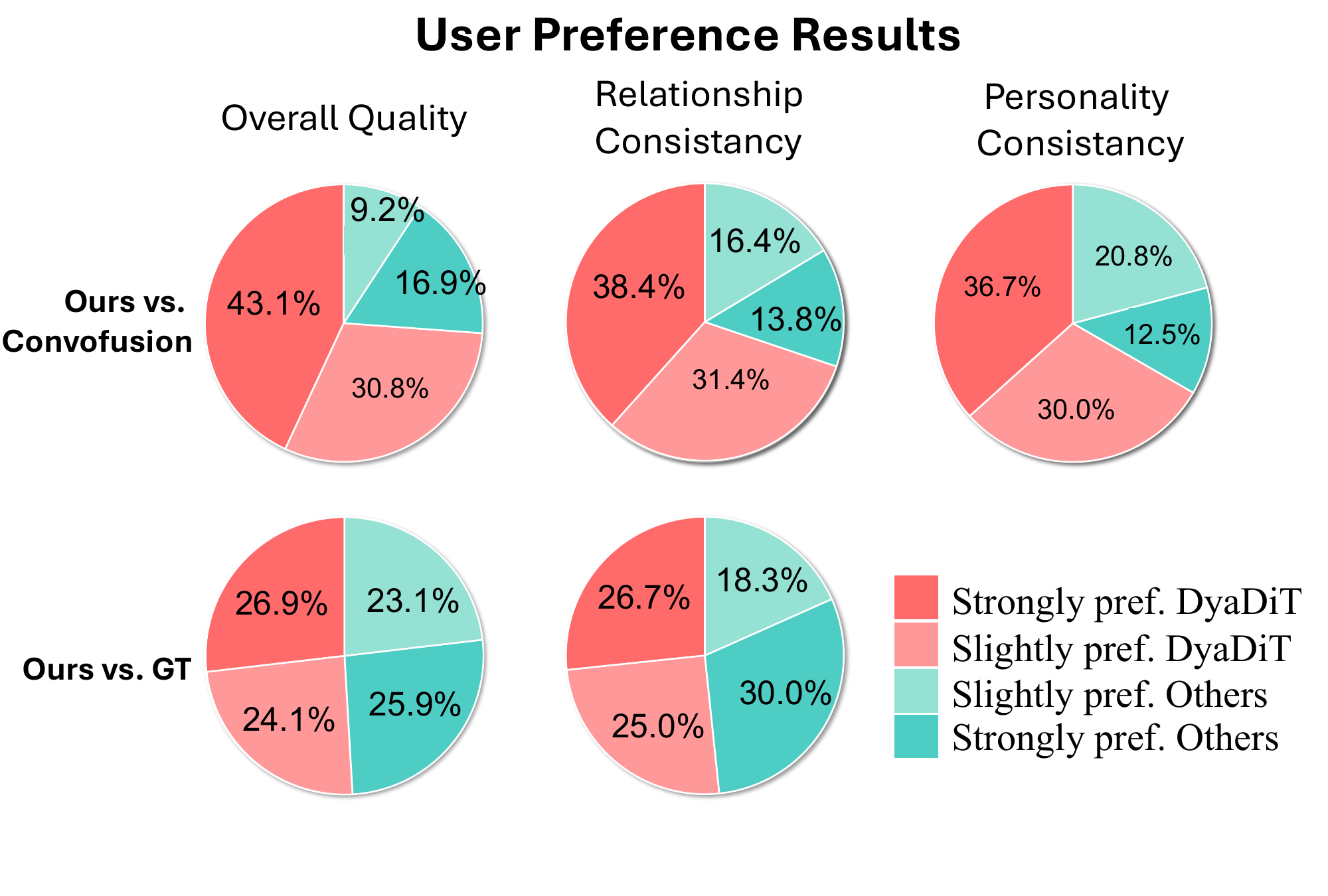}
    \caption{A/B subjective evaluation percentages comparing our method with ConvoFusion\cite{convofusion2024} and with ground truth. Participants preferred our generated motion due to its more natural and socially aware conversational behavior.}
    \label{fig:pie}
    \vspace{-0.5cm}
\end{figure}

As shown in Figure~\ref{fig:pie}, DyaDiT achieves notably higher user preference scores than \textit{ConvoFusion}~\cite{convofusion2024}, and interestingly, even slightly outperforms the ground truth gestures from the Seamless Interaction dataset. 


Specifically, DyaDiT is preferred by 73.9\%, 69.8\%, and 66.7\% of participants in overall gesture quality, relationship consistency, and personality consistency, respectively, showing that it generates both higher-quality and more socially appropriate gestures. Each participant evaluated 56 video pairs (10\,s each), requiring about 20-25 minutes in total, and each pair was rated by 16 participants (aged 25-35, with CS backgrounds). Both quality and personality show strong deviations from chance against the baseline ($\chi^2$, $p<10^{-8}$), while personality also remains significantly preferred over the ground truth ($\chi^2$, $p=1.5\times10^{-4}$). Our generated gestures are further preferred over the ground truth by 1.0\% and 1.7\% in two settings, likely because diffusion produces smoother motions and social conditioning encourages slightly more expressive gestures. We do not compare against the ground truth in personality consistency, since personality annotations are continuous and difficult for participants to judge reliably.

\section{Conclusion}
In this work, we present \textbf{DyaDiT}, a multi-modal dyadic gesture generation model designed to produce socially consistent and natural conversational behaviors between two speakers. By jointly conditioning on dyadic audio, social attributes, and the partner's motion, DyaDiT effectively captures interpersonal dynamics in dyadic conversation. Our orthogonalization cross-attention (ORCA) module helps clarify the contribution of each speaker’s audio, and the motion dictionary enhances expressive richness by providing style-aware motion priors. Both quantitative and subjective evaluations demonstrate that DyaDiT achieves superior performance compared to existing methods in terms of realism, diversity, and social coherence. We believe this work takes an important step toward socially favorable synthetic agents and opens up future research directions, such as socially aware dual-agent gesture generation.
\section{Limitation \& Future Work}
\label{sec:limitation}

Currently, our model is limited by the dataset and may still suffer from conditioning conflicts, as personality cues can be implicitly encoded in the audio, reducing controllability and diversity. In future work, we plan to explore audio neutralization to better disentangle social cues, collect extensive dyadic gesture datasets using our existing human reconstruction pipeline~\cite{songcontact4d, shinbodycontact4d} for full-body motion generation.

\section{Acknowledgment}
This work was supported by JST ASPIRE JPMJAP2404, and JST CRONOS JPMJCS24N8.
\vspace{-1.4cm}

{
    \small
    \bibliographystyle{ieeenat_fullname}
    \bibliography{main}
}

\clearpage
\setcounter{page}{1}
\maketitlesupplementary

This supplementary material contains sections below:
\begin{itemize}
    \item \textbf{1.} Relationships \& Personality Clustering of Generated Results
    \item \textbf{2.} Implementation Details of DyaDiT
    \item \textbf{3.} Details of A/B Test Questionnaire
\end{itemize}

In addition to this \texttt{supplementary.pdf}, we also include a 
\texttt{narration\_video.mp4}, which provides a brief overview of the paper along with several qualitative gesture generation examples. 
We further provide our implementation in \texttt{dyadit\_code.zip}; please refer to the included \texttt{README.md} for instructions on running the code. 
The trained models will be released upon acceptance.

\section{Clustering of Generated Gestures}
\label{sec:clustering}
In the main paper, we conduct an A/B test to evaluate the relationship and personality consistency of the generated gestures. To further assess the controllability of the conditional inputs in DyaDiT, we perform a t-SNE clustering analysis on the generated motion embeddings.

\begin{figure}[h!]
    \centering
    \includegraphics[width=\linewidth]{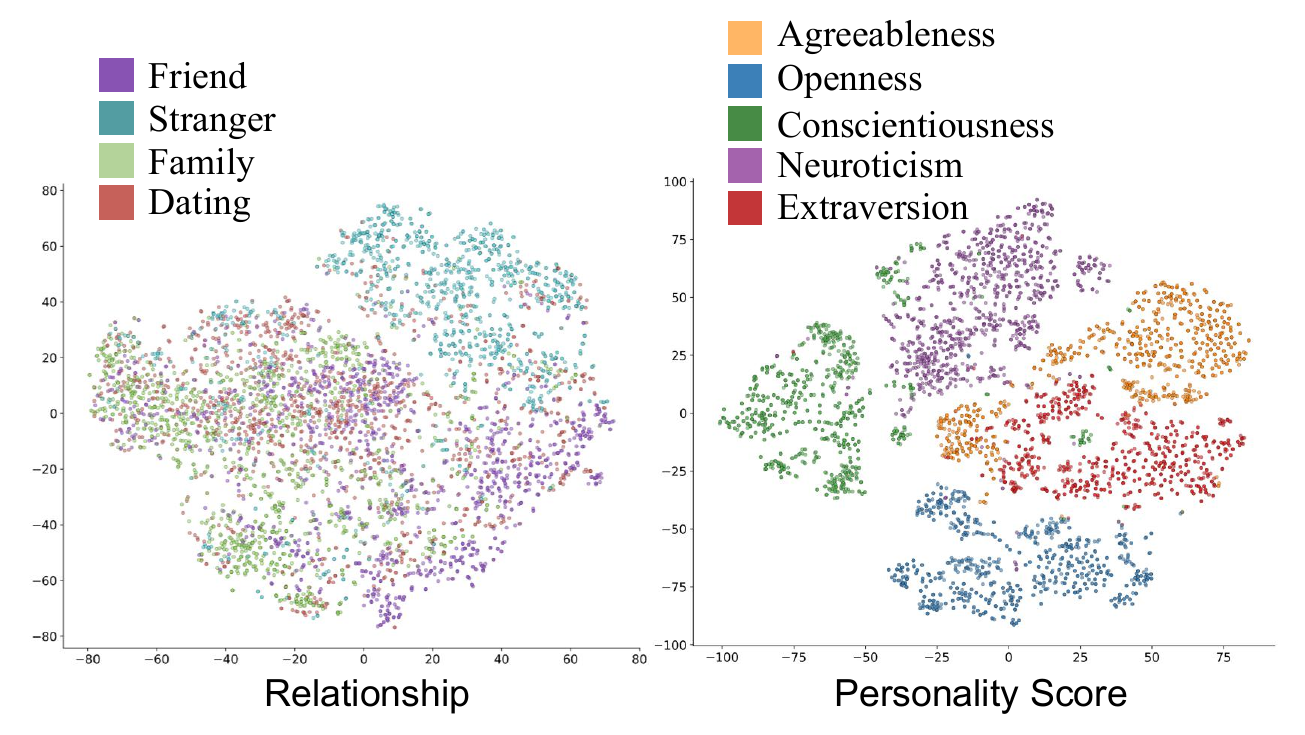}
    \caption{t-SNE clustering results of Relationships (left), Personality Scores (right). }
    \label{fig:tsne-sup}
\end{figure}

Figure~\ref{fig:tsne-sup} visualizes the t-SNE embeddings of the generated gestures under different conditioning signals.  
On the left, we generate gestures using various \textit{relationship} types while fixing the personality scores.  
On the right, we discretize the continuous \textit{personality score} features into five ``one-hot'' vectors and generate gestures for each vector to examine personality controllability.

We observe that the personality clusters form clearly separable groups, indicating that DyaDiT effectively captures the global behavioral tendencies associated with different personality traits.  
In contrast, the relationship clusters are less clearly separated. We consider this to be consistent with the nature of dyadic conversational gestures: the styles between \textit{Friend}, \textit{Family}, and \textit{Dating} share some overlap. As a result, the generated gestures also show a more continuous manifold across these categories rather than sharp cluster boundaries.

\section{Implementation Details}
\label{sec:implementation}
\paragraph{Diffusion Transformer}
The input pose sequence is encoded into a latent space aligned with the VQ-VAE representation, resulting in a latent embedding in $\mathbb{R}^{64}$, instead of the original $\mathbb{R}^{43\times6}$ 6D rotation matrix~\cite{zhou2018rot6d} joint representation.
A linear layer projects the noisy latent input from $\mathbb{R}^{64}$ to a hidden space in $\mathbb{R}^{512}$, followed by a symmetric projection back to $\mathbb{R}^{64}$ at the output.

The model contains 4 Transformer blocks, each equipped with 4-head multi-head attention ($\mathbb{R}^{128}$) and a $\mathbb{R}^{2048}$ feed-forward network.
We employ a $\mathbb{R}^{512}$ sinusoidal time embedding, which is injected into each block through FiLM~\cite{film2018} modulation.

Two independent Wav2Vec2~\cite{wav2vec2} processors extract high-level audio features from the conversational speech of both speakers.
These yield feature sequences denoted as $a_{\text{self}}, a_{\text{other}} \in \mathbb{R}^{T \times 768}$, where $T$ denotes the number of audio frames.
Each of these is projected to $\mathbb{R}^{T \times 512}$ via a linear transformation, followed by LayerNorm and a gated fusion mechanism to combine self and other speaker cues.

A learnable motion bank contains 1000 prototype vectors in $\mathbb{R}^{512}$, providing contextual priors via cross-attention.
similar to the time embedding, relationship and personality embeddings are projected into $\mathbb{R}^{512}$.
These vectors are injected into the DiT blocks via FiLM-style adaptive scaling.
All contextual cues are concatenated into a unified sequence in $\mathbb{R}^{512}$ and injected into each DiT block via cross-attention.

\paragraph{Motion Tokenizer (VQ-VAE).}
We implement a temporal VQ-VAE~\cite{lee2022autoregressiveimagegenerationusing} to discretize pose sequences before diffusion.
Given an input sequence of joint features $X \in \mathbb{R}^{T \times 6 \times 43}$, the encoder is a 1D CNN consisting of three Conv1d layers, with LeakyReLU applied after the first two layers, and an overall temporal downsampling factor of 4, producing a latent sequence in $\mathbb{R}^{(T/4) \times 64}$.
This continuous latent is quantized by a residual vector quantizer with depth 4, each equipped with a 512-entry codebook, which maps each time step to a stack of discrete code indices.
The decoder is a 1D CNN consisting of an initial Conv1d-LeakyReLU layer, two upsampling blocks (linear interpolation followed by Conv1d and LeakyReLU) interleaved with additional Conv1d-LeakyReLU refinement layers, and a final Conv1d projection, achieving an overall temporal upsampling factor of 4 and recovering the original temporal resolution to reconstruct poses in $\mathbb{R}^{T \times 6 \times 43}$.
The final latent representation used by the DiT denoiser is obtained from the quantized codes as a compact 64-dimensional embedding per $4 \times T$ frames in $\mathbb{R}^{64}$.

\paragraph{Seamless Interactive Dataset.}
We conduct our experiments on a part of the Seamless Interaction dataset~\cite{seamless_interaction}. In particular, we adopt the \textit{naturalistic} split of the dataset. For training, we utilize the first 10 official training archives (provided as zip files), which contain approximately 182 hours of naturalistic interactions and 3000 paired motion-audio samples. For testing, we select the first archive from the official test split to ensure a consistent evaluation setting.

We observe that the SMPL-H parameters provided in the dataset exhibit noticeable inaccuracies in lower body estimation, likely due to limited camera views and body occlusions during data capture. To avoid introducing artifacts into our motion modeling, we discard lower-body joints and only retain the upper body comprising 43 joints, including fingers. For visualization, all unused joints, along with global orientation and root translation, are set to zero.

In addition to pose data, the dataset includes high level annotations such as \textit{relationship} and \textit{personality scores}. While the dataset provides \textit{Interpersonal Communicative Dynamic} (IPC) tags for social dynamics, we found the annotations to be too noisy and ambiguous where it is unclear which speaker they apply to. Consequently, we do not employ IPC tag supervision in our current study and instead focus on the cleaner relationship and personality cues. We note that once the IPC annotations are refined in future dataset releases, we plan to extend our framework with an IPC-awared conditioning module to further capture communicative intent in dyadic gestures.

In the future, we plan to re-annotate the video data using advanced human pose estimation tools such as SMPLest-x~\cite{yin2025smplest}, Harmony4D~\cite{khirodkar2024harmony4d} or recent state-of-the-art models, with the aim of obtaining more reliable full body motion supervision.

\section{Questionnaire}

We provide a reconstructed version of the A/B test questionnaire used in our user study. 
To view the questionnaire, please first extract the \texttt{questionnaire\_video.zip} file inside the 
\texttt{questionnaire} folder. After extraction, open \texttt{questionnaire.html} in any modern web browser.

\begin{figure}[h!]
    \centering
    \includegraphics[trim={1.3cm 0cm 1cm 0cm}, clip, width=1.0\linewidth]{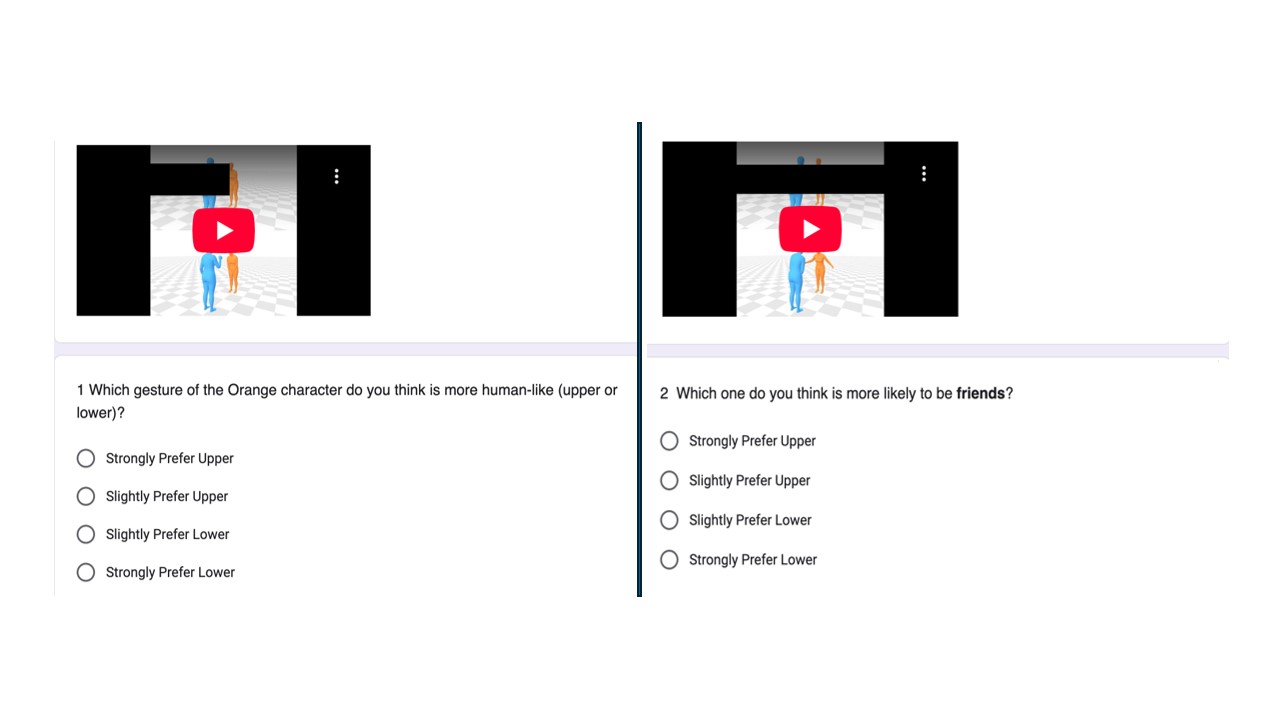}
    \caption{Example of Questionnaires in GoogleForm}
    \label{fig:GoogleForm}
\end{figure}

The original questionnaire was conducted through Google Forms (see Figure~\ref{fig:GoogleForm}). It consists of 
$28 \times 2$ questions in total, including 10 questions on overall gesture quality, 
8 questions on relationship consistency, and 10 questions on personality consistency. 
Each question presents paired gesture videos for comparison under two settings: 
\textit{DyaDiT vs.\ ConvoFusion} and \textit{DyaDiT vs.\ Ground Truth}.

For an accurate viewing experience, please wear headphones.  
The left audio channel corresponds to the partner’s speech, while the right audio channel 
corresponds to the target speaker’s speech. 

The reconstructed interface allows reviewers to browse all questions and play the 
corresponding videos to experience the same evaluation procedure as our participants.

\end{document}